\documentclass[letterpaper,10pt,conference]{ieeeconf}  

\IEEEoverridecommandlockouts                              

\usepackage{mymacros}
\usepackage{times} 
\usepackage{graphicx}

\usepackage[T1]{fontenc}
\usepackage{color}
\usepackage{hyperref}
\usepackage{pifont}

\usepackage{caption} 

\let\labelindent\relax
\usepackage{enumitem}

\title{\LARGE \bf
    Affordance RAG: Hierarchical Multimodal Retrieval with Affordance-Aware Embodied Memory for Mobile Manipulation
}

\author{
    Ryosuke Korekata$^{1,2,3}$, Quanting Xie$^{3}$, Yonatan Bisk$^{3}$, and Komei Sugiura$^{1,2}$
\thanks{
    $^{1}$Keio University, $^{2}$Keio AI Research Center, $^{3}$Carnegie Mellon University.
    {\tt\small rkorekata@keio.jp}
}
\thanks{
    This work was partially supported by by Microsoft Corporation as part of the Keio CMU partnership, JSPS Fellows Grant Number JP25KJ2068, JSPS KAKENHI Grant Number 23K03478, and JST Moonshot.
}
}

\begin{document}

\makeatletter
\let\@oldmaketitle\@maketitle 
\renewcommand{\@maketitle}{\@oldmaketitle 
}
\makeatother

\newcommand{\shiftDocumentDownWithMargins}[1]{%
  \addtolength{\topmargin}{#1}      
  \addtolength{\textheight}{-#1}   
  \addtolength{\footskip}{#1}      
  \addtolength{\textheight}{#1}    
}

\maketitle
\thispagestyle{empty}
\pagestyle{empty}
\shiftDocumentDownWithMargins{3mm} 


\begin{abstract}
    In this study, we address the problem of open-vocabulary mobile manipulation, where a robot is required to carry a wide range of objects to receptacles based on free-form natural language instructions.
    This task is challenging, as it involves understanding visual semantics and the affordance of manipulation actions.
    To tackle these challenges, we propose Affordance RAG, a zero-shot hierarchical multimodal retrieval framework that constructs Affordance-Aware Embodied Memory from pre-explored images.
    The model retrieves candidate targets based on regional and visual semantics and reranks them with affordance scores, allowing the robot to identify manipulation options that are likely to be executable in real-world environments.
    Our method outperformed existing approaches in retrieval performance for mobile manipulation instruction in large-scale indoor environments.
    Furthermore, in real-world experiments where the robot performed mobile manipulation in indoor environments based on free-form instructions, the proposed method achieved a task success rate of 85\%, outperforming existing methods in both retrieval performance and overall task success.
 \end{abstract}


\section{Introduction}

As robots are increasingly deployed in real-world human environments, such as homes, hospitals, and warehouses, there is a growing demand for systems that can understand and execute flexible, language-driven instructions.
The goal of our work is open-vocabulary mobile manipulation (OVMM~\cite{Yenamandra2023HomeRobotOM}) guided, where a robot is required to identify and interact with objects and receptacles described in free-form language.
Given an instruction and a set of pre-explored environment images, the robot must retrieve the appropriate target object and receptacle, that it can successfully manipulate  (e.g., pick and place) in the real-world.

A typical use case of our target problem is a domestic service robot instructed with a natural language instruction such as ``Please bring the paper towels to the kitchen counter.''
To execute this instruction, the robot must first identify the target object and the receptacle from a set of previously observed images of the environment.
Furthermore, when multiple paper towels are present in the environment, the robot is expected to select the one with a higher grasping affordance.
Similarly, within the kitchen counter, it should prefer an area that is uncluttered and more suitable for object placement.
This task is challenging due to the need for open-vocabulary grounding and affordance-aware reasoning.

A naive approach would involve directly applying a vision-language model (VLM) to evaluate every candidate image against the instruction.
However, this becomes computationally impractical in realistic settings, where hundreds of candidate views or object instances must be considered per scene.
Most existing approaches (e.g.,~\cite{Yenamandra2023HomeRobotOM,liu2024okrobot,liu2024dynamem}) and multimodal retrieval methods (e.g.,~\cite{clip,longclip,blip2,beit3}), tend to misidentify visually similar but semantically incorrect candidates (e.g., confusing a lotion bottle with a water bottle).
Crucially, these methods lack affordance awareness---retrieving objects that are physically non-manipulable, resulting in downstream execution failures.


\begin{figure}
    \centering
    \includegraphics[clip,width=\linewidth]{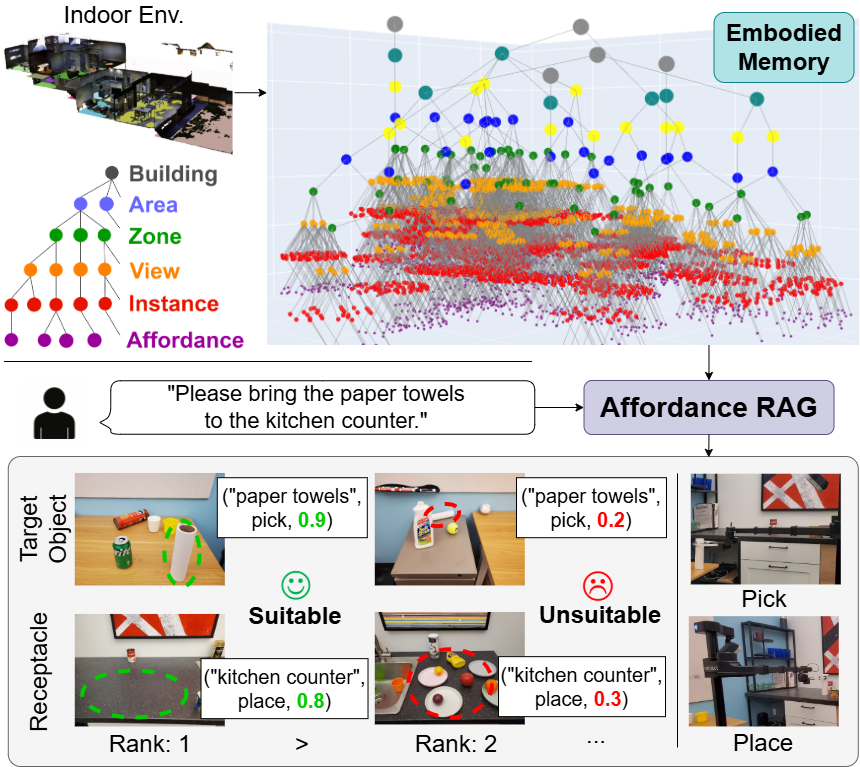}
    \caption{
        \small
        Overview of Affordance RAG for open-vocabulary mobile manipulation.
        The robot first constructs an embodied memory based on images collected during pre-exploration of the environment.
        When a free-form instruction is given, hierarchical multimodal retrieval is performed over the embodied memory to identify the target object and receptacle.
        To improve task success rates, candidates with higher affordance scores are prioritized during retrieval.
    }
    \label{fig:eyecatch}
    \vspace{-5mm}
\end{figure}

To address the limitations, we propose Affordance RAG, a hierarchical multimodal retrieval framework that integrates multi-level semantic representations with robotic affordance-aware reasoning.
Fig.~\ref{fig:eyecatch} shows an overview of the proposed method.
The main difference between our method and prior approaches lies in its hierarchical multimodal retrieval framework that fuses regional and visual semantics, and its ability to incorporate affordance-aware reasoning through affordance-centric memory and reranking.
While existing methods typically perform flat similarity matching, our method constructs a structured Affordance-Aware Embodied Memory (Affordance Mem) and refines multimodal retrieval through VLM-based affordance estimation and large language model (LLM)-guided object selection, enabling robust zero-shot mobile manipulation.
We introduce Affordance-Aware Reranking to address a fundamental limitation in existing retrieval-based approaches: their inability to distinguish between semantically relevant candidates and those more suitable for execution.
Our Affordance-Aware Reranking module filters candidates using VLM-predicted affordances and scores instance-level descriptions with an LLM for relevance, followed by reranking top candidates based on affordance scores to achieve both linguistic consistency and suitability.

The key contributions of this work are three-fold:
\begin{itemize}
    \setlength{\parskip}{0mm}
    \setlength{\itemsep}{0.2mm}
    \item We propose Affordance RAG, a zero-shot hierarchical multimodal retrieval framework that combines regional and visual semantics via Multi-Level Fusion.
    \item We introduce Affordance Mem by using instance-level Affordance Proposer based on VLMs via visual prompting to estimate robot affordances.
    \item We introduce Affordance-Aware Reranking that combines affordance prefiltering with LLM-based descriptive instance retrieval, and reranks candidates based on affordance scores to achieve both linguistic relevance and suitability.
\end{itemize}

\vspace{-2mm}
\section{Related Work}
\vspace{-1mm}
\subsection{Language-Guided Mobile Manipulation}
\vspace{-0.5mm}

Mobile manipulation tasks based on natural language instructions have been widely studied, with several real-world benchmarks (e.g.,~\cite{Yenamandra2023HomeRobotOM}).
While these benchmarks rely on template-based instructions, our work addresses the more challenging task of free-form, OVMM.
The OVMM task has been extensively studied in recent work~\cite{liu2024okrobot,liu2024dynamem,korekata23iros}.
Approaches that construct 3D scene graphs to represent the environment have been proposed for mobile manipulation tasks (e.g.,~\cite{momallm24}).
Beyond this, scene graphs have also been applied to embodied question answering (EQA~\cite{yang20243dmem3dscenememory,Xie2024EmbodiedRAGGN}), navigation~\cite{osg,Werby-RSS-24,wang2025navraggeneratinguserdemand}, task planning~\cite{gu2024conceptgraphs}, and semantic mapping~\cite{hughes2022hydra}.
These approaches are constructed based on object detection or semantic segmentation, but rely heavily on object category labels, making them less suitable for the free-form OVMM task considered in this work.

\vspace{-1mm}
\subsection{Multimodal Retrieval}
\vspace{-0.5mm}

Text-image retrieval using multimodal foundation models has been extensively studied~\cite{beit3, blip2,longclip,clip}.
Recent work has actively explored applying these models to robotics, where a robot is given natural language instructions and retrieves target objects from pre-explored environment images to execute manipulation tasks~\cite{nlmap, Sigurdsson2023RRExBoTRR,relaxformer, dm2rm}.
In addition, several studies have explored applying the concept of retrieval-augmented generation (RAG) to robotics, including approaches that construct real-world environments as hierarchical embodied memories (e.g., Embodied-RAG~\cite{Xie2024EmbodiedRAGGN}, \cite{wang2025navraggeneratinguserdemand}).
Embodied-RAG is closely related to our work, but differs in that it focuses on what is visible in the scene without explicitly considering robot affordances.
In contrast, our approach models high-level robot affordances grounded in atomic actions.
Affordance modeling for object manipulation has been explored at different levels, including predicting object functionality from 3D point clouds~\cite{delitzas2024scenefun3d}, estimating contact points from images~\cite{bahl2023affordances}, and representing affordances as high-level atomic actions~\cite{jiang2024roboexp}, as in our work.

\subsection{Foundation Models for Robotics}

Foundation models as multimodal AI agents hold great promise for a wide range of applications in scene understanding and planning (e.g.,~\cite{song2025robospatial}).
A growing body of research has explored applying LLMs and VLMs to robotic tasks~\cite{hu2023toward}.
For example, LLMs and VLMs have been applied to commonsense reasoning (e.g.,~\cite{Driess2023palme}) and task planning~\cite{brohan2023saycan}.
Several prior works~\cite{yang2023arxiv,relaxformer} enhance the reasoning capabilities of VLMs by applying visual prompts to input images.
In contrast, our method uses visual prompting to generate object-centric descriptions and predict robot-executable affordances from observed images.

\section{Problem Statement}

\begin{figure*}[t]
    \centering
    \includegraphics[clip,width=\linewidth]{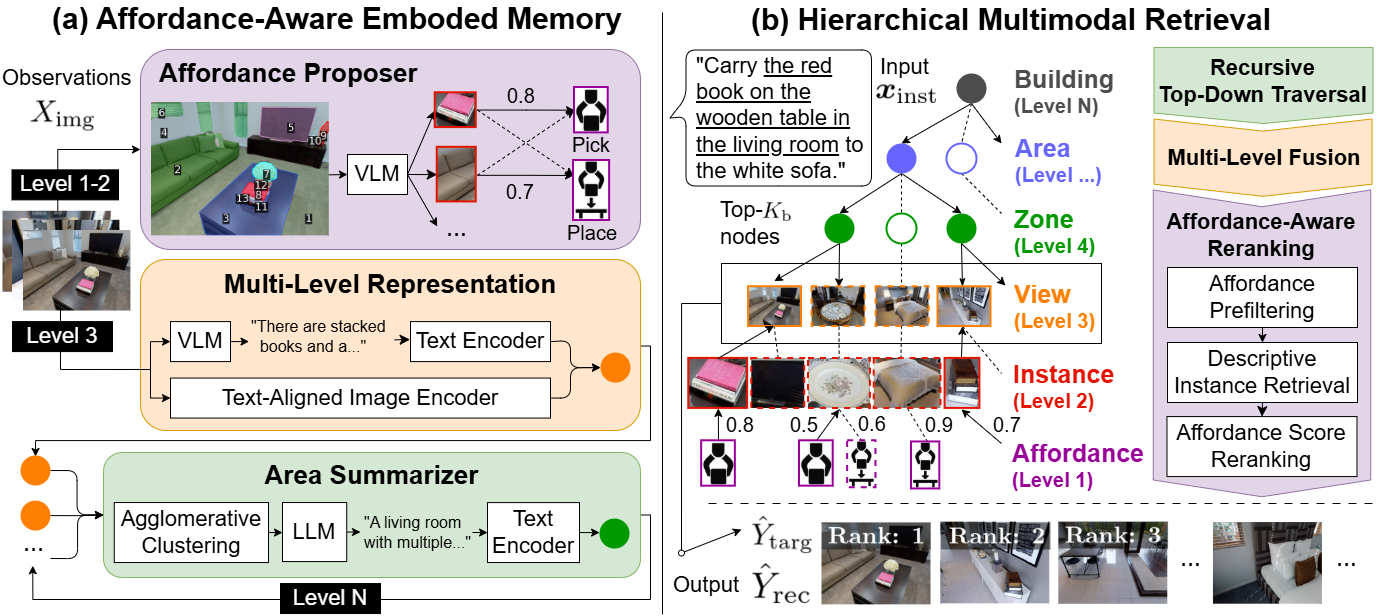}
    \caption{
        \small
        Overview of the Affordance RAG framework.
        (a) The robot constructs Affordance-Aware Embodied Memory (Affordance Mem) through pre-exploration.
        Affordance Mem is constructed from three components: Affordance Proposer, Multi-Level Representation, and Area Summarizer.
        (b) Upon receiving an instruction, hierarchical multimodal retrieval is performed to identify both the target object and the receptacle.
        This process consists of three stages: Recursive Top-Down Traversal, Multi-Level Fusion, and Affordance-Aware Reranking.
    }
    \label{fig:framework}
    \vspace{-3mm}
\end{figure*}

In this study, we focus on the Multimodal Retrieval-guided Mobile Manipulation (MRMM) task.
In this task, given a natural language instruction, a robot performs mobile manipulation by retrieving the images of the target object and the receptacle from a set of environmental images.
The target object and the receptacle are identified based on the user's selections from the top-retrieved images.
This task comprises two sub-tasks: \textit{multimodal retrieval} and \textit{action execution}.
During the multimodal retrieval phase, it is expected that the images of the target object and the receptacle are ranked highly in their respective retrieved image lists.
During the action execution phase, the robot is expected to pick up the target object and transport it to the designated receptacle.
The input is defined as $\bm{x} = \left\{\bm{x}_\mathrm{inst}, X_\mathrm{img}\right\}$, $X_\mathrm{img} = \left\{\bm{x}_\mathrm{img}^{(j)}\right\}_{j=1}^{N_\mathrm{img}}$, where $\bm{x}_\mathrm{inst}$ and $\bm{x}_\mathrm{img}^{(j)} \in \mathbb{R}^{3 \times W \times H}$ denote an instruction and an RGB image with width $W$ and height $H$, respectively.
Here, $N_\mathrm{img}$ denotes the number of candidate images.
The output consists of two ranked lists of images corresponding to the target object and the receptacle, respectively.

The terminology used in this paper is defined as follows:
The ``target object'' and ``receptacle'' are the everyday object and the desired receptacle (piece of furniture), specified in the instruction.
%
We assume that images of the environment have been collected through pre-exploration.
This is a realistic setting because mobile service robots are typically deployed to perform tasks repeatedly in the same known environment (e.g., \cite{Sigurdsson2023RRExBoTRR,relaxformer}).

\section{Proposed Method}

We propose Affordance RAG, a zero-shot hierarchical multimodal retrieval framework for OVMM.
Fig.~\ref{fig:framework} shows the overview of the proposed method.
To understand what kind of robot affordances are executable in an environment, we introduce Affordance-Aware Embodied Memory (Affordance Mem), constructed from observed images through pre-exploration, as shown in Fig~\ref{fig:framework} (a).
Our method utilizes Affordance Mem as a real-world database for RAG to identify both the target object and the receptacle, as shown in Fig.~\ref{fig:framework} (b).
While our primary focus is on OVMM, the proposed Affordance Mem is considered to be broadly applicable to other embodied reasoning tasks such as EQA~\cite{Majumdar_2024_CVPR,yang20243dmem3dscenememory}.

\subsection{Affordance-Aware Embodied Memory}

For robust OVMM, it is important to identify the most suitable option in terms of robot affordance when multiple valid candidates exist.
However, existing embodied memories (e.g.,~\cite{Xie2024EmbodiedRAGGN,wang2025navraggeneratinguserdemand}) and scene graphs (e.g.,~\cite{momallm24,Werby-RSS-24}) focus solely on what is present in the scene.
To address this limitation, we propose a hierarchical embodied memory that spans multiple levels of nodes---from robot affordance, instance, view, zone, and area to building---across levels 1 through $N$.
As illustrated in Fig.~\ref{fig:framework} (a), Affordance Mem is constructed sequentially in a bottom-up manner by obtaining nodes through the following three components: (1) Affordance Proposal, which predicts instance-level affordances from visual observations; (2) Multi-Level Representation, which obtains view-level regional and visual semantics; and (3) Area Summarization, which aggregates multi-view features to form regional nodes.

\subsubsection{\textbf{Affordance Proposer (Level 1-2)}}

In this module, we apply object-centric visual prompting to $\bm{x}_\mathrm{img}^{(j)}$ and feed the resulting image into VLMs (e.g., GPT-4o~\cite{gpt_4o}), to extract instance-level representations and robot affordances, which are stored as levels 2 and 1 nodes, respectively.
This enables the extraction of descriptive instance-level expressions (e.g., ``red metal mug,'' ``antique wooden side table with vertical slats'') that go beyond category-level nodes (e.g., ``cup,'' ``table'') typically constructed via object detection or semantic segmentation~\cite{momallm24,Werby-RSS-24}.
Specifically, we use SEEM~\cite{zou24neurips} to generate a segmentation mask for $\bm{x}_\mathrm{img}^{(j)}$, and feed both the original image and a visual prompt image---where segmented regions are overlaid with indexed numbers---into the VLM in parallel.
Although previous studies~\cite{yang2023arxiv,relaxformer} have proposed a similar captioning enhancement approach, our work extends this idea to affordance-aware memory construction.
The output is a set of instance-affordance triplets $\mathcal{A}^{(j)} = \left\{(\bm{o}_k, a_k, f_k)\right\}_{k=1}^{N_\mathrm{af}}$, where $\bm{o}_k$, $a_k$, $f_k$, and $N_\mathrm{af}$ denote the instance representation, the type of robot affordance, the affordance score, and the number of such triplets, respectively.
We focus on mobile manipulation tasks and consider ``pick'' and ``place'' as the primary robot affordances.
However, as the Affordance Proposer is a prompt-based module, it can be easily extended to other atomic actions such as ``open'' and ``close'' without any architectural modification.

\subsubsection{\textbf{Multi-Level Representation (Level 3)}}

In this module, we construct a Multi-Level Representation for $\bm{x}_\mathrm{img}^{(j)}$ by explicitly combining two types of features---high-level regional and low-level visual semantics---which are stored as a level 3 node.
Specifically, the former refers to $\bm{l}^{(3,j)}_\mathrm{s} \in \mathbb{R}^{d_\mathrm{t}}$, obtained by embedding the image description generated by VLMs using a text encoder (e.g., text-embedding-3-large~\cite{txtembedding3large}).
The latter, $\bm{v}^{(3,j)} \in \mathbb{R}^{d_\mathrm{m}}$, is obtained using a text-aligned image encoder from a multimodal foundation model (e.g., BEiT-3~\cite{beit3}).
Here, $d_\mathrm{t}$ and $d_\mathrm{m}$ denote the dimensionality of the respective embeddings.
The fusion strategy for combining these features is detailed in Sec.~\ref{sec:mlf}.
The generated image descriptions are also used in the Area Summarizer module to perform hierarchical node aggregation.

\subsubsection{\textbf{Area Summarizer (Level $N$)}}

In this module, we apply agglomerative clustering to generate level $(i+1)$ nodes by aggregating nodes at level $i$ for $i \ge 3$.
Most existing multimodal foundation models~\cite{beit3,blip2,longclip,clip} handle each image independently as a feature vector, making it challenging to capture out-of-view contexts.
However, in our target task, instructions often contain referring expressions related to out-of-view objects or room-level semantics (e.g., ``... towel near the sink...,'' ``... in the bedroom...''), which should be considered during retrieval.
Therefore, this module performs agglomerative clustering based on both the physical and semantic distances between nodes.

In this module, we repeat the following procedure recursively up to level $N$: Initially, clustering is performed on the basis of the Euclidean distance between node positions and the cosine similarity between their textual descriptions in the embedding space.
Subsequently, for each cluster, we generate a summary of the node descriptions using an LLM, and embed the resulting text into $\bm{l}^{(i+1,j)}_\mathrm{s} \in \mathbb{R}^{d_\mathrm{t}}$ using the text encoder.
This hierarchical representation is expected to enable regional and semantically grounded understanding of the environment during retrieval.

\subsection{
    Hierarchical Multimodal Retrieval
}

Fig.~\ref{fig:framework} (b) illustrates the procedure for performing hierarchical multimodal retrieval over Affordance Mem given a natural language instruction.
The hierarchical multimodal retrieval process is performed sequentially in a top-down manner, comprising the following three stages: (1) Recursive Top-Down Traversal, which explores the hierarchical memory to progressively narrow down candidate regions; (2) Multi-Level Fusion, which ranks view-level nodes by integrating regional and visual semantics; and (3) Affordance-Aware Reranking, which refines the top-retrieved nodes by evaluating affordance.

\subsubsection{\textbf{Recursive Top-Down Traversal}}

In this module, we recursively traverse the memory from level $N$ down to level 3 by selecting the top-$K_\mathrm{b}$ nodes at each level based on the similarity score $s^{(i,j)} = \mathrm{sim}(\bm{l}_\mathrm{t}, \bm{l}^{(i,j)}_\mathrm{s})$, where $\mathrm{sim}(\cdot, \cdot)$, $\bm{l}_\mathrm{t} \in \mathbb{R}^{d_\mathrm{t}}$, and $\bm{l}^{(i,j)}_\mathrm{s} \in \mathbb{R}^{d_\mathrm{t}}$ denote cosine similarity, the text embedding of the instruction, and the text embedding of the $j$-th node at level $i$, respectively.
This hierarchical traversal enables coarse-to-fine filtering based on regional semantics such as areas and zones.
Moreover, this module selects appropriate nodes by embedding similarity because previous approaches~\cite{Xie2024EmbodiedRAGGN,wang2025navraggeneratinguserdemand}, which utilize LLM for selecting nodes, often struggle with global selection from a large number of candidates, presenting a key limitation.
Since node embeddings can be precomputed and stored during pre-exploration, this also enables faster inference than those existing methods.
As the output of this module, a set of view-level (level 3) nodes $\mathcal{S} \subseteq \{ (\bm{v}^{(3,j)}, \bm{l}_s^{(3,j)}) \}$ is extracted.

\subsubsection{
    \textbf{Multi-Level Fusion}
    \label{sec:mlf}
}

This module performs complementary multimodal retrieval using the high-level regional and low-level visual semantics obtained from the Multi-Level Representation.
Specifically, for each element in $\mathcal{S}$, we compute the similarity score $s_\text{mlf}^{(3,j)}$ as follows:
\begin{align*}
    s_\text{mlf}^{(3,j)} = \alpha \cdot \mathrm{sim}(\bm{l}_\mathrm{t}, \bm{l}^{(3,j)}_\mathrm{s}) + (1 - \alpha) \cdot \mathrm{sim}(\bm{l}_\mathrm{m}, \bm{v}^{(3,j)}),
\end{align*}
where $\alpha \in [0,1]$ and $\bm{l}_\mathrm{m} \in \mathbb{R}^{d_\mathrm{m}}$ denote a weighting hyperparameter and a feature vector obtained from the text encoder of a multimodal foundation model (e.g., BEiT-3), respectively.
This represents a weighted sum of regional semantics obtained via hierarchical traversal and visual semantics derived from a multimodal foundation model.
Prior work has typically utilized either the former (e.g.,~\cite{Xie2024EmbodiedRAGGN,wang2025navraggeneratinguserdemand}) or the latter (e.g.,~\cite{beit3,blip2,longclip,clip}) in isolation.
By fusing these regional semantics with visual semantics, Multi-Level Fusion enables complementary multimodal retrieval that considers both global contextual consistency and fine-grained visual similarity.

\subsubsection{\textbf{Affordance-Aware Reranking}}

This module reranks the top-$K_\mathrm{r}$ nodes in $\mathcal{S}$---initially sorted by Multi-Level Fusion---using $\{\mathcal{A}^{(j)}\}$ from levels 1 and 2.
The reranking process consists of three steps: Affordance Prefiltering, Descriptive Instance Retrieval, and Affordance Score Reranking (ASR).
In the first stage, we prefilter instance nodes based on robot affordances: ``pick'' for target object image retrieval and ``place'' for receptacle image retrieval.
This allows the method to narrow down candidates based on both object appearance and functional perspective. 
In the second stage, we provide the descriptive expressions generated by the Affordance Proposer to an LLM, which scores the filtered instance nodes based on their similarity to the instruction.
While LLMs are less effective at global retrieval, they excel at matching within a small set of candidates, making this setting well-suited to their strengths.
In the third stage, the top-$K_\mathrm{f}$ instance nodes are further refined by reranking them based on their affordance scores.
Finally, we perform fine-grained reranking over the top-$K_\mathrm{r}$ view nodes by prioritizing those that contain the instance nodes selected in this module.
This hierarchical multimodal retrieval process is executed separately for the target object and the receptacle, resulting in ranked image lists $\hat{Y}_\mathrm{targ}$ and $\hat{Y}_\mathrm{rec}$, respectively.

\section{Experiments}

\begin{table*}[t]
    \centering
    \normalsize
    \caption{
        \small
        Quantitative comparison between the proposed method and baseline methods on the WholeHouse-MM benchmark. 
        The best and second-best scores for each metric are indicated in \textbf{bold} and \underline{underline}, respectively.
        ``*'' denotes reproduced results.
        }
    \vspace{-1mm}
    \resizebox{\textwidth}{!}{
    \begin{tabular}{lccccccccccccc}
    \toprule
    {[\%]}
    & \multicolumn{3}{c}{Target Object}  
    & \multicolumn{3}{c}{Receptapcle}  
    & \multicolumn{3}{c}{Overall}  
    \\ 
    \cmidrule(lr){2-4}
    \cmidrule(lr){5-7}
    \cmidrule(lr){8-10}
    Method
    & \small R@5 $\uparrow$
    & \small R@10 $\uparrow$
    & \small R@20 $\uparrow$
    & \small R@5 $\uparrow$
    & \small R@10 $\uparrow$
    & \small R@20 $\uparrow$
    & \small R@5 $\uparrow$
    & \small R@10 $\uparrow$
    & \small R@20 $\uparrow$
    
    \\ 
    \midrule 
    CLIP \cite{clip}
    & {15.7}		
    & {24.2} 
    & {33.6} 
    & {6.2} 		
    & {11.7} 
    & {21.7}
    & {10.9}
    & {18.0}
    & {27.7}
    \\

    Long-CLIP \cite{longclip}
    & {24.6} 
    & {36.1} 
    & {48.5} 
    & {3.5} 
    & {9.2} 
    & {19.9} 
    & {14.0}
    & {22.6}
    & {34.2}
    \\


    
    BLIP-2 \cite{blip2}
    & {\underline{30.2}} 
    & {40.7} 
    & {48.0} 
    & {5.2} 
    & {10.3} 
    & {19.9} 
    & {17.7}
    & {25.5}
    & {34.0}
    \\
    
    BEiT-3 \cite{beit3}
    & {29.0} 
    & {\underline{42.1}}
    & {\underline{53.8}} 
    & {\underline{8.9}} 
    & {15.4} 
    & {27.6}
    & {\underline{19.0}}
    & {\underline{28.7}}
    & {\underline{40.7}}
    \\
    
    \midrule
    
    HomeRobot* \cite{Yenamandra2023HomeRobotOM}
    & {5.9} 
    & {10.2} 
    & {12.9} 
    & {1.7} 
    & {3.9} 
    & {8.4} 
    & {3.8}
    & {7.0}
    & {10.7}
    \\
    
    NLMap* \cite{nlmap}
    & {15.1} 
    & {19.2} 
    & {30.4} 
    & {7.3} 
    & {15.1} 
    & {23.6} 
    & {11.2} 
    & {17.2} 
    & {27.0}
    \\
    
    RelaX-Former \cite{relaxformer}
    & {17.9} 
    & {26.5} 
    & {37.7} 
    & {8.8} 
    & {\underline{19.8}} 
    & {\underline{28.4}} 
    & {13.3}
    & {23.2}
    & {33.0}
    \\
    
    
    
    
    
    \midrule
    
    Embodied-RAG \cite{Xie2024EmbodiedRAGGN} 
    & {15.1} 
    & {18.5} 
    & {22.8} 
    & {6.7} 
    & {11.3} 
    & {14.4}
    & {10.9}
    & {14.9}
    & {18.6} 
    \\
    
    \textbf{Affordance RAG (ours)}
    & \textbf{32.8}
    & \textbf{49.9}
    & \textbf{61.7}
    & \textbf{14.8}
    & \textbf{24.3}
    & \textbf{30.2}
    & \textbf{23.8}
    & \textbf{37.1}
    & \textbf{45.9}
    
    \\ \bottomrule
    \end{tabular}
    }
    \label{tab:simulated}
    \vspace{-2mm}
\end{table*}

\subsection{WholeHouse-MM Benchmark}

We introduce the WholeHouse-MM benchmark, constructed from the Matterport3D (MP3D~\cite{chang2017matterport3d}) dataset.
In this task, building-scale indoor environments with hundreds of images and human-annotated (not generated) instructions for mobile manipulation are required.
Most existing benchmarks addressing open-vocabulary mobile manipulation either use template-based instructions~\cite{Yenamandra2023HomeRobotOM} or are not designed for building-scale task execution~\cite{relaxformer,dm2rm}.
Therefore, we collected images from MP3D, a standard dataset widely used for research on navigation and scene understanding in indoor environments, enabling the evaluation of multimodal retrieval at the building scale.
While \cite{qi2020reverie,zhu2021soon} also focus on referring expression comprehension tasks in MP3D environments, they do not handle mobile manipulation instructions.
Thus, following \cite{relaxformer,dm2rm}, the WholeHouse-MM benchmark uses human-annotated instructions containing referring expressions collected via crowdsourcing.

To collect images from each environment, we simulated a pre-exploration phase in MP3D.
Since MP3D provided a map of the environment, we captured panoramic views by rotating the camera at each waypoint in 60-degree increments, collecting six images per location.
Each environment contained an average of 590 images.
The instructions in WholeHouse-MM were collected via crowdsourcing from 116 annotators.
The annotators were presented with two images from the environment---one depicting the target object and the other the receptacle---and asked to give instructions for carrying the target object to the receptacle.
The target objects and receptacles were obtained by extracting the locations of predefined object categories from REVERIE~\cite{qi2020reverie}, a standard benchmark for Vision-and-Language Navigation tasks.
If a target object or receptacle appeared in multiple viewpoints, those images were also treated as positive.

The benchmark consists of 402 instructions and 2,360 images collected from real-world indoor environments.
The vocabulary size is 517, with a total of 6,410 words and an average sentence length of 15.9 words.
The environments for each split were selected according to \cite{relaxformer,dm2rm}.
The validation set was used for hyperparameter tuning, while the test set was used for evaluating the performance of the methods.

\vspace{-1mm}
\subsection{Quantitative Results}

Table~\ref{tab:simulated} shows the quantitative comparison between the proposed method and baseline methods.
Since the mobile manipulation instructions include both a target object and a receptacle, we reported multimodal retrieval performance for each component as well as the overall score.
Note that in this benchmark, we focus on retrieval performance and omit physical manipulation; therefore, the ASR step was excluded from the proposed method.
For results that include this step, please see Sec.~\ref{sec:physical}.
We used recall@$K$ ($K$ = 5, 10, 20) as the evaluation metric.
The primary evaluation metric was recall@10.
We used recall@$K$ as it is a standard evaluation metric in image retrieval settings~\cite{liu2009learning}.
We used eight baseline methods: CLIP (ViT-L/14)~\cite{clip}, Long-CLIP (ViT-L/14)~\cite{longclip}, BLIP-2 (ViT-g)~\cite{blip2}, BEiT-3 (large)~\cite{beit3}, HomeRobot~\cite{Yenamandra2023HomeRobotOM}, NLMap~\cite{nlmap}, RelaX-Former~\cite{relaxformer}, and Embodied-RAG~\cite{Xie2024EmbodiedRAGGN}.
Except for RelaX-Former, all methods were evaluated in a zero-shot setting.

\begin{figure}[t]
    \centering
    \includegraphics[width=0.92\linewidth]{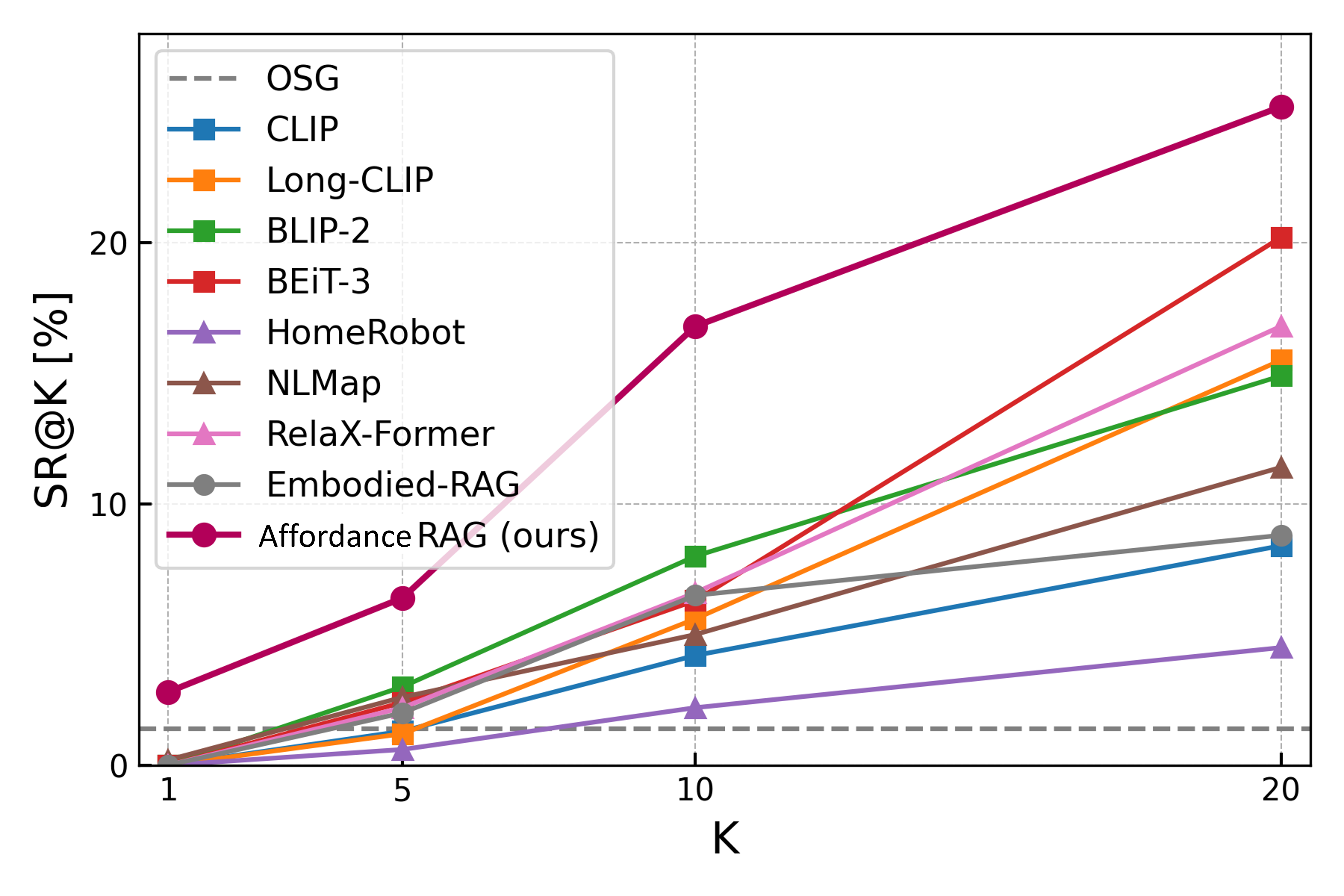}
    \vspace{-4mm}
    \caption{
        \small
        Comparison of task success rate (SR) on the WholeHouse-MM benchmark. SR@$K$ denotes the percentage of samples in which both the target object and the receptacle are correctly retrieved within the top-$K$ results.
    }
    \label{fig:simulated_sr}
    \vspace{-3mm}
\end{figure}

\begin{table*}[t]
\centering
\small
\caption{
    \small
    Results of ablation studies on the WholeHouse-MM benchmark.
    The best scores for each metric are indicated in \textbf{bold}.
}
\vspace{-2mm}
\resizebox{\textwidth}{!}{
\begin{tabular}{lccccccccccc}
\toprule
{[\%]}
& Regional
& Visual
& Affordance-Aware
& Affordance
& \multicolumn{3}{c}{Overall} 
\vspace{-1mm}

\\

\cmidrule(lr){6-8}

Method

& Semantics
& Semantics
& Reranking
& Proposer
& \small R@5 $\uparrow$
& \small R@10 $\uparrow$
& \small R@20 $\uparrow$

\\ 
\midrule


(a)
& 
& $\checkmark$
& $\checkmark$
& $\checkmark$

& {22.9}
& {32.1}
& {40.7}
\\

(b)
& $\checkmark$
& 
& $\checkmark$
& $\checkmark$
& {20.3}
& {30.0}
& {39.8}

\\

(c)
& $\checkmark$
& $\checkmark$
& 
& $\checkmark$

& {20.1}
& {29.5}
& {41.5}
\\

(d)
& $\checkmark$
& $\checkmark$
& $\checkmark$
& 

& {20.0}
& {31.2}
& {41.5}

\\

(e)
& $\checkmark$
& $\checkmark$
& $\checkmark$
& $\checkmark$
& \textbf{23.8}
& \textbf{37.1}
& \textbf{45.9}

\\ \bottomrule
\end{tabular}
}
\label{tab:ablation}
\vspace{-5mm}
\end{table*}

As shown in Table~\ref{tab:simulated}, our proposed method achieved recall@10 scores of 49.9\%, 24.3\%, and 37.1\% for the target object, receptacle, and overall metrics, respectively.
In contrast, the best-performing baseline method achieved 42.1\%, 19.8\%, and 28.7\%, indicating that our method outperformed it by 7.8, 4.5, and 8.4 points, respectively.
Similarly, our method outperformed the best baseline method across all evaluation metrics.

Furthermore, to compare our approach with the Object Goal Navigation method that also utilizes a hierarchical memory (OSG~\cite{osg}), we present the task success rate (SR) comparison results in Fig.~\ref{fig:simulated_sr}.
SR@$K$ denotes the percentage of samples in which both the target object and the receptacle are correctly retrieved within the top-$K$ results.
As shown in Fig.~\ref{fig:simulated_sr}, OSG obtained an SR of 1.4\%, whereas our proposed method achieved 2.8\%, 6.4\%, 16.8\%, and 25.2\% for $K = 1, 5, 10, 20$, respectively, the best performance among all methods.

\subsection{Qualitative Results}

\begin{figure}[t]
    \centering
    \includegraphics[clip,width=\linewidth]{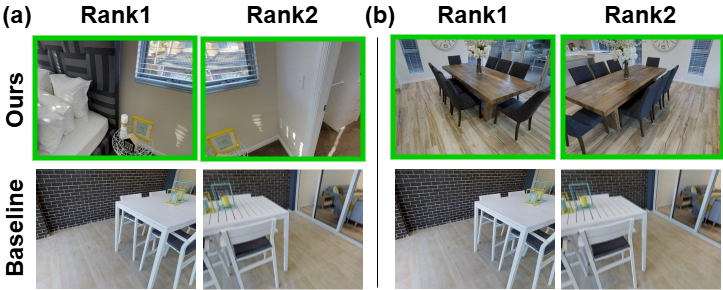}
    \vspace{-3mm}
    \caption{
        \small
        Qualitative results on the WholeHouse-MM benchmark.
        The given $\bm{x}_\mathrm{inst}$ was ``Take a photo frame from the side table in the bedroom and place it on the dining table with a bouquet of flowers.''
        (a) Target object and (b) receptacle: Top-2 retrieved images are shown for both the our method and the best baseline method (BEiT-3~\cite{beit3}).
        The ground-truth image is highlighted with a \textcolor[rgb]{0.2,0.8,0}{green} border.
    }
    \label{fig:qualitative_simulated}
    \vspace{-3mm}
\end{figure}

Fig.~\ref{fig:qualitative_simulated} shows a successful example from the WholeHouse-MM benchmark.
In this example, $\bm{x}_\mathrm{inst}$ was ``Take a photo frame from the side table in the bedroom and place it on the dining table with a bouquet of flowers.''
As shown in Fig.~\ref{fig:qualitative_simulated} (a), the baseline method ranked an unrelated white table among the top candidates, whereas the proposed method successfully ranked the photo frame placed on the side table in the bedroom as the top-1 and top-2 candidates.
Similarly, as shown in Fig.~\ref{fig:qualitative_simulated} (b), the proposed method ranked the dining table with a bouquet of flowers as both the top-1 and top-2 candidates.
These results suggest that the proposed hierarchical multimodal retrieval method, which incorporates both regional and visual semantics, is effective.

\vspace{-1mm}    
\subsection{Ablation Studies}

To validate our framework, we conducted ablation studies on the construction of Affordance Mem and the hierarchical multimodal retrieval strategy.
Table~\ref{tab:ablation} shows the ablation results on the WholeHouse-MM benchmark.

\textbf{Multi-Level Fusion ablation:}\;
To investigate the effectiveness of the node representation, we conducted ablation studies by removing the features related to regional and visual semantics.
Specifically, (a) for the former, we excluded $\bm{l}^{(3,j)}_\mathrm{s}$, which was obtained via hierarchical retrieval; and (b) for the latter, we excluded $\bm{v}^{(3,j)}$, derived from the text-aligned image encoder.
As shown in Table~\ref{tab:ablation}, Methods (a) and (b) achieved recall@10 scores of 32.1\% and 30.0\%, respectively, which were 5.0 and 7.1 points lower than Method (e).
These results support our design of Multi-Level Fusion, where high-level semantic reasoning (regional semantics) and low-level perceptual grounding (visual semantics) jointly contribute to robust instruction-grounded retrieval.

\textbf{Affordance-Aware Reranking ablation:}\;
To investigate the effectiveness of reranking using instance- and affordance-level nodes in Affordance Mem, we evaluated a variant without the reranking step.
As shown in Table~\ref{tab:ablation}, Method (c) achieved a recall@10 of 29.5\%, which was 7.6 points lower than Method (e).
This result suggests that reranking based on instance- and affordance-level nodes in Affordance Mem enables more instruction-consistent multimodal retrieval.

\textbf{Affordance Proposer ablation:}\;
To validate the effectiveness of descriptive affordance proposals generated via visual prompting, we implemented a method that performed LLM-based reranking using only image captions generated by VLMs.
Table~\ref{tab:ablation} shows that Method (d) achieved a recall@10 of 31.2\%, which was 5.9 points lower than Method (e).
This result suggests that beyond relying solely on VLM-generated captions, incorporating structured information about nodes is beneficial for reranking.

\begin{figure}[t!]
    \centering
    \includegraphics[width=0.9\linewidth]{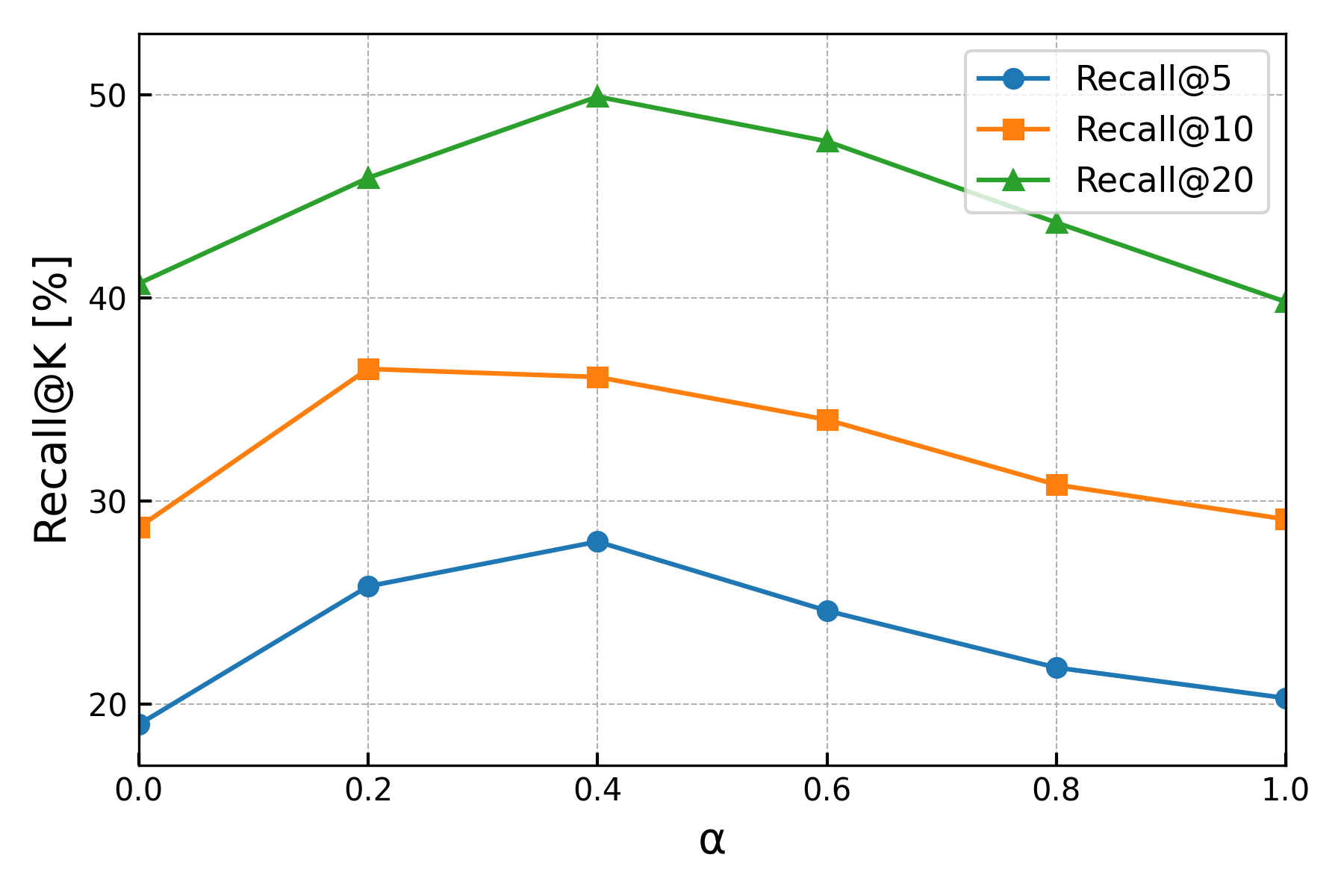}
    \vspace{-4mm}
    \caption{
        \small
        Sensitivity analysis of the weight $\alpha$ in Multi-Level Fusion. The hyperparameter $\alpha$ balances the contribution between regional semantics and visual semantics.
    }
    \label{fig:sensitivity}
    \vspace{-3mm}
\end{figure}

\textbf{Impact of Multi-Level Fusion weight $\alpha$:}\;
Fig.~\ref{fig:sensitivity} shows the sensitivity analysis of the weight $\alpha$ in Multi-Level Fusion.
We investigated the effect of $\alpha$ on recall@$K$ by varying it from 0.0 to 1.0 in increments of 0.2.
The results show that performance degrades when the retrieval was biased toward either regional semantics or visual semantics.
This suggests that the two representations play complementary roles in multimodal retrieval, and appropriately balancing them leads to a more comprehensive understanding.

\section{
    Real-World Experiments
    \label{sec:physical}
}

We validated our method through real-world experiments using a mobile manipulator.
In particular, we aimed to evaluate the effectiveness of ASR on the task success rate by executing pick-and-place actions in the real world, where the suitability of object grasping and placement varies across objects.

\subsection{Settings \& Implementation}

We used a Hello Robot Stretch 2~\cite{kemp2022design} equipped with a DexWrist for mobile manipulation.
This robot is commonly used as a standard platform for OVMM tasks~\cite{Yenamandra2023HomeRobotOM,liu2024dynamem,liu2024okrobot}.
The environment used in our experiments was a $5.0 \times 7.0$ m$^2$ indoor room consisting of office and kitchen areas, containing 10 different pieces of furniture.
We used the 20 everyday objects as target objects.
Among them, 14 objects were randomly selected from the YCB objects~\cite{calli15ram}, which are widely used in manipulation research, and the remaining 6 were composed of commonly used household objects.
In our experiments, we assumed that these objects were initially placed on randomly selected pieces of furniture.


Firstly, to construct Affordance Mem, the robot first performed a pre-exploration phase.
During this phase, the robot captured RGB images using an Intel RealSense D435i camera from a pre-defined viewpoint that allowed observation of the entire environment.
Path planning and navigation followed standard map-based approaches.
The robot constructed a 2D map using Hector SLAM~\cite{kohlbrecher2011flexible}.
Next, the user provided a free-form instruction.
The user were asked to provide an instruction that required transporting a randomly selected object in the environment to a randomly selected piece of furniture.
We conducted 40 trials in total, using a different instruction for each trial.

After receiving the instruction, the robot performed the following steps.
First, the robot retrieved images of the target object and receptacle from Affordance Mem and presented the top-5 retrieved images for each to the user.
In real-world experiments, when performing affordance prediction based on a VLM, we provided a prompt that included information about the robot's embodiment (e.g., gripper shape and width, arm length).
The inference took 0.12 seconds per instruction and used 16 GB of VRAM on an NVIDIA Geforce RTX 3090.
Next, the robot navigated to the location where the user-selected target object image had been captured, and performed the grasping action.
The grasping point was determined as the median of the point cloud corresponding to the segmented mask of the target object, obtained from the depth image and the segmentation produced by SAM~\cite{kirillov23iccv}.
Similarly, the robot transported the target object to the location where the user-selected receptacle image had been captured and placed the target object.
Since motion generation for grasping, placement, and navigation is out of the scope of this study, we adopted heuristic-based methods.

\vspace{-2mm}
\subsection{Quantitative Results}

\begin{table}
    \centering
    \small
    \caption{
        \small
        Quantitative comparison in the real-world experiments.
        The best scores for each metric are indicated in \textbf{bold}.
        The scores are the results from 40 trials.
        }
    \vspace{-1mm}
    \resizebox{\columnwidth}{!}{
        \begin{tabular}{lcccccc}
            \toprule
            
            Method {[\%]}
            & \small R@5 $\uparrow$
            & \small \text{SR} $\uparrow$
            \\ 
            
            \midrule 
            
            BEiT-3~\cite{beit3}
            & {79} 
            & {45}
            \\

            Affordance RAG (w/o ASR)
            & \textbf{94}
            & 70
            \\

            \textbf{Affordance RAG (full)}
            & \textbf{94}
            & \textbf{85}
            \\

            \bottomrule
        \end{tabular}
    }
    \label{tab:physical}
    \vspace{-4mm}
\end{table}



Table~\ref{tab:physical} shows the quantitative results of the real-world experiments.
We used recall@5 and SR as evaluation metrics in our experiments.
A trial was considered successful only if the method retrieved a correct image for both the target object and the receptacle within the top-5 results, and the robot executed the pick and place actions without failure.
Recall@5 indicates the overall score for both the target object and the receptacle.
%
We used BEiT-3~\cite{beit3} as a baseline method because it was the best-performing baseline in the simulated experiments, as shown in Table~\ref{tab:simulated}.

As shown in Table~\ref{tab:physical}, the proposed method achieved a recall@5 of 94\%, outperforming the baseline method by 15 points.
Similarly, the proposed method achieved an SR of 85\%, which is 40 points higher than the baseline method.
These results demonstrate that the proposed method can be successfully integrated into a real-world robot to enable open-vocabulary mobile manipulation.
The recall@5 scores in Table~\ref{tab:simulated} are generally lower than those in Table~\ref{tab:physical} due to the larger building-scale search space of MP3D.

To validate the effectiveness of ASR in the proposed method, we conducted an ablation experiment by removing this step.
As shown in Table~\ref{tab:physical}, the proposed method achieved an SR of 85\%, compared to 70\% without ASR, demonstrating a 15-point improvement.
These results suggest that when instructions are ambiguous and allow for multiple valid interpretations, ASR helps prioritize candidates with higher pick-and-place suitability, thereby improving SR.
The identical recall@5 scores for both methods are due to reranking being applied only within the top-5 retrieved candidates.

\begin{figure}[t]
    \centering
    \includegraphics[clip,width=\linewidth]{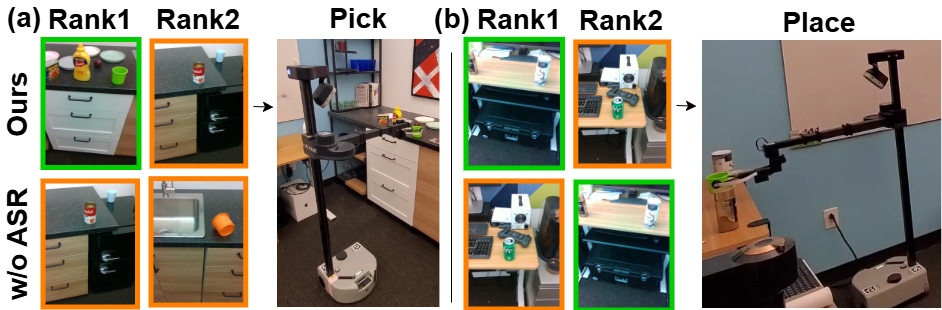}
    \caption{
        \small
        Qualitative results of the real-world experiments.
        The given $\bm{x}_\mathrm{inst}$ was ``Please deliver a cup to the desk that has some coffee powder on it.''
        (a) Target object and (b) receptacle: Top-2 retrieved images and real-world execution results are shown for both the proposed method and the variant without Affordance Score Reranking (ASR).
        The ground-truth image and semantically correct but less suitable image are highlighted with \textcolor[rgb]{0.2,0.8,0}{green} and \textcolor[RGB]{252,140,3}{orange} borders, respectively.
    }
    \label{fig:qualitative_physical}
    \vspace{-5mm}
\end{figure}

\vspace{-1.5mm}
\subsection{Qualitative Results}

Fig.~\ref{fig:qualitative_physical} shows a successful example from the real-world experiments.
In this example, $\bm{x}_\mathrm{inst}$ was ``Please deliver a cup to the desk that has some coffee powder on it.''
As shown in the results, our proposed method ranked a green cup placed upright on the kitchen counter as the top candidate due to its high suitability, while the variant without ASR ranked less suitable options higher, such as a blue cup placed deep on top of the refrigerator and an orange cup lying on its side.
Similarly, the receptacle was a desk with coffee powder on it; however, two desks were present in the environment, separated by a coffee machine.
According to the results, the proposed method ranked the tidier desk with fewer objects as the top candidate, while the variant without ASR ranked the more cluttered desk higher.
Based on the retrieval results, the robot was able to grasp the green cup and successfully transport it to the desk with coffee powder.
This suggests that ASR improved SR by promoting the selection of more suitable objects in response to ambiguous instructions.



\section{Conclusions}

In this study, we address the task of MRMM, where a robot performs mobile manipulation based on language instructions by retrieving target object and receptacle images from environmental images.
We proposed Affordance RAG, a zero-shot hierarchical multimodal retrieval framework that combines regional and visual semantics via Multi-Level Fusion based on Affordance Mem.
Affordance RAG outperformed the baseline methods in terms of standard metrics on the newly built WholeHouse-MM benchmark.
Furthermore, in real-world experiments, the proposed method achieved a task success rate of 85\%, outperforming existing methods in both retrieval performance and overall task success.
One limitation of the proposed method is that all relevant visual information is assumed to be covered by the observed images, which may not hold in highly occluded scenes.
In addition, our current framework focuses on visual and spatial affordance reasoning and does not explicitly consider physical or kinematic constraints of robot actions.
As future work, we plan to address both limitations by combining our approach with active exploration and physical reasoning.




\bibliographystyle{IEEEtran}
\bibliography{reference}

@inproceedings{kohlbrecher2011flexible,
  title={{A Flexible and Scalable SLAM System with Full 3D Motion Estimation}},
  author={Kohlbrecher, Stefan and Von Stryk, Oskar and Meyer, Johannes and Klingauf, Uwe},
  booktitle={SSRR},
  pages={155--160},
  year={2011}
}

@article{osg,
  title={{Open Scene Graphs for Open World Object-Goal Navigation}},
  author={Loo, Joel and Wu, Zhanxin and Hsu, David},
  journal={IJRR},
  year={2025}
}

@inproceedings{kemp2022design,
  title={{The Design of Stretch: A Compact, Lightweight Mobile Manipulator for Indoor Human Environments}},
  author={Kemp, Charles and Edsinger, Aaron and Clever, Henry and Matulevich, Blaine},
  booktitle={ICRA},
  pages={3150--3157},
  year={2022},
}

@InProceedings{kirillov23iccv,
author    = {Kirillov, Alexander and Mintun, Eric and Ravi, Nikhila and Mao, Hanzi and Rolland, Chloe and Gustafson, Laura and Xiao, Tete and Whitehead, Spencer and Berg, Alexander and Lo, Wan and Dollar, Piotr and Girshick, Ross},
title     = {{Segment Anything}},
booktitle = {ICCV},
year      = {2023},
pages     = {4015--4026}
}

@ARTICLE{calli15ram,
  author={Calli, Berk and Walsman, Aaron and Singh, Arjun and Srinivasa, Siddhartha  and others},
  journal={IEEE RAM}, 
  title={{Benchmarking in Manipulation Research: Using the Yale-CMU-Berkeley Object and Model Set}}, 
  year={2015},
  volume={22},
  number={3},
  pages={36-52},
}

@inproceedings{brohan2023saycan,
  title={{Do As I Can, Not As I Say: Grounding Language in Robotic Affordances}},
  author={Brohan, Anthony and Chebotar, Yevgen and Finn, Chelsea and Hausman, Karol and Herzog, Alexander and Ho, Daniel and Ibarz, Julian and Irpan, Alex and others},
  booktitle={CoRL},
  pages={287--318},
  year={2023}
}

@inproceedings{Driess2023palme,
title = {{PaLM-E: An Embodied Multimodal Language Model}},
author = {Driess, Danny and Xia, Fei and Sajjadi, Mehdi and Lynch, Corey and Chowdhery, Aakanksha and Ichter, Brian and Wahid, Ayzaan and Tompson, Jonathan and Vuong, Quan and others},
booktitle = {ICML},
year = {2023},
pages={8469--8488},
}

@article{hu2023toward,
  title={{Toward General-Purpose Robots via Foundation Models: A Survey and Meta-Analysis}},
  author={Hu, Yafei and Xie, Quanting and Jain, Vidhi and Francis, Jonathan and Patrikar, Jay and Keetha, Nikhil and Kim, Seungchan and Xie, Yaqi and others},
  journal={arXiv preprint arXiv:2312.08782},
  year={2023}
}

@inproceedings{song2025robospatial,
  author    = {Song, Chan and Blukis, Valts and Tremblay, Jonathan and Tyree, Stephen and Su, Yu and Birchfield, Stan},
  title     = {{{RoboSpatial}: Teaching Spatial Understanding to {2D} and {3D} Vision-Language Models for Robotics}},
  booktitle = {CVPR},
  pages={15768--15780},
  year      = {2025},
}

@inproceedings{jiang2024roboexp,
      title={{RoboEXP: Action-Conditioned Scene Graph via Interactive Exploration for Robotic Manipulation}},
      author={Jiang, Hanxiao and Huang, Binghao and Wu, Ruihai and Li, Zhuoran and Garg, Shubham and Nayyeri, Hooshang and Wang, Shenlong and Li, Yunzhu},
      booktitle={CoRL},
      year={2024}
}

@inproceedings{bahl2023affordances,
  title={{Affordances from Human Videos as a Versatile Representation for Robotics}},
  author={Bahl, Shikhar and Mendonca, Russell and Chen, Lili and Jain, Unnat and Pathak, Deepak},
  booktitle={CVPR},
  pages={13778--13790},
  year={2023}
}

@inproceedings{delitzas2024scenefun3d, 
  title = {{SceneFun3D: Fine-Grained Functionality and Affordance Understanding in 3D Scenes}}, 
  author = {Delitzas, Alexandros and Takmaz, Ayca and Tombari, Federico and Sumner, Robert and others}, 
  booktitle = {CVPR}, 
  pages={14531--14542},
  year = {2024}
}

@inproceedings{hughes2022hydra,
    title={{Hydra: A Real-time Spatial Perception System for {3D} Scene Graph Construction and Optimization}},
    author={N. Hughes and Y. Chang and L. Carlone},
    booktitle={RSS},
    year={2022},
}

@inproceedings{gu2024conceptgraphs,
  title={{ConceptGraphs: Open-Vocabulary 3D Scene Graphs for Perception and Planning}},
  author={Gu, Qiao and Kuwajerwala, Ali and Morin, Sacha and Jatavallabhula, Krishna and Sen, Bipasha and Agarwal, Aditya and Rivera, Corban and others},
  booktitle={ICRA},
  pages={5021--5028},
  year={2024},
}

@INPROCEEDINGS{korekata23iros,
  author={Korekata, Ryosuke and Kambara, Motonari and Yoshida, Yu and Ishikawa, Shintaro and others},
  booktitle={IROS}, 
  title={{Switching Head-Tail Funnel UNITER for Dual Referring Expression Comprehension with Fetch-and-Carry Tasks}}, 
  year={2023},
  pages={3865-3872},
}

@article{liu2024dynamem,
  title={{DynaMem: Online Dynamic Spatio-Semantic Memory for Open World Mobile Manipulation}},
  author={Liu, Peiqi and Guo, Zhanqiu and Warke, Mohit and Chintala, Soumith and Paxton, Chris and others},
  journal={arXiv preprint arXiv:2411.04999},
  year={2024}
}

@article{liu2024okrobot,
  title={{OK-Robot: What Really Matters in Integrating Open-Knowledge Models for Robotics}},
  author={Liu, Peiqi and Orru, Yaswanth and Paxton, Chris and Shafiullah, Nur Muhammad Mahi and Pinto, Lerrel},
  journal={arXiv preprint arXiv:2401.12202},
  year={2024}
}

@misc{txtembedding3large,
    author={{OpenAI}},
    title={{text-embedding-3-large}},
    year={2024},
    howpublished={\url{https://platform.openai.com/docs/models/embeddings}},
    note={Accessed: Jun. 2025}
}

@article{yang2023arxiv,
title={{Set-of-Mark Prompting Unleashes Extraordinary Visual Grounding in GPT-4V}},
author={Yang, Jianwei and Zhang, Hao and Li, Feng and Zou, Xueyan and Li, Chunyuan and Gao, Jianfeng},
journal={arXiv preprint arXiv:2310.11441},
year={2023}
}

@inproceedings{zou24neurips,
title={{Segment Everything Everywhere All at Once}},
author={Zou, Xueyan and Yang, Jianwei and Zhang, Hao and Li, Feng and Li, Linjie and Wang, Jianfeng and Wang, Lijuan and Gao, Jianfeng and Lee, Yong},
booktitle={NeurIPS},
pages = {19769-19782},
year={2023}
}

@misc{gpt_4o,
  author       = {OpenAI},
  title        = {{GPT-4o: Optimized Generative Pre-trained Transformer 4}},
  year         = {2024},
  howpublished = {\url{https://openai.com}},
  note         = {Accessed: Jun. 2025}
}

@INPROCEEDINGS{Werby-RSS-24, 
              AUTHOR    = {Abdelrhman Werby AND Chenguang Huang AND Martin Büchner AND Abhinav Valada AND Wolfram Burgard}, 
              TITLE     = {{Hierarchical Open-Vocabulary 3D Scene Graphs for Language-Grounded Robot Navigation}}, 
              BOOKTITLE = {RSS}, 
              YEAR      = {2024}, 
          }

@article{momallm24,
  title={{Language-Grounded Dynamic Scene Graphs for Interactive Object Search with Mobile Manipulation}},
  author={Daniel Honerkamp and Martin Büchner and Fabien Despinoy and others},
  journal={IEEE RA-L},
  volume={9},
  number={10},
  pages={2377-3766},
  year={2024}
}

@InProceedings{yang20243dmem3dscenememory,
      title={{3D-Mem: 3D Scene Memory for Embodied Exploration and Reasoning}}, 
      author={Yuncong Yang and Han Yang and Jiachen Zhou and Peihao Chen and Hongxin Zhang and Yilun Du and Chuang Gan},
      booktitle = {CVPR},
      pages={17294--17303},
      year={2025},
}

@InProceedings{Majumdar_2024_CVPR,
    author    = {Majumdar, Arjun and Ajay, Anurag and Zhang, Xiaohan and Putta, Pranav and Yenamandra, Sriram and Henaff, Mikael and Silwal, Sneha and Mcvay, Paul and others},
    title     = {{OpenEQA: Embodied Question Answering in the Era of Foundation Models}},
    booktitle = {CVPR},
    year      = {2024},
    pages     = {16488-16498}
}

@article{liu2009learning,
  title={{Learning to Rank for Information Retrieval}},
  author={Liu, Tie},
  journal={FNTIR},
  volume={3},
  number={3},
  pages={225--331},
  year={2009},
}

@inproceedings{zhu2021soon,
  title={{SOON: Scenario Oriented Object Navigation with Graph-based Exploration}},
  author={Zhu, Fengda and Liang, Xiwen and Zhu, Yi and Yu, Qizhi and Chang, Xiaojun and Liang, Xiaodan},
  booktitle={CVPR},
  pages={12689--12699},
  year={2021}
}

@INPROCEEDINGS{qi2020reverie,
  author={Qi, Yuankai and Wu, Qi and Anderson, Peter and Wang, Xin and Wang, William and Shen, Chunhua and Hengel, Anton},
  booktitle={CVPR}, 
  title={{REVERIE: Remote Embodied Visual Referring Expression in Real Indoor Environments}}, 
  year={2020},
  pages={9979--9988},
}

@article{dm2rm,
  author    = {R. Korekata and others},
  title     = {{DM\textsuperscript{2}RM: Dual-Mode Multimodal Ranking for Target Objects and Receptacles Based on Open-Vocabulary Instructions}},
  journal   = {Advanced Robotics},
  volume    = {39},
  number    = {5},
  pages     = {243--258},
  year      = {2025}
}

@inproceedings{chang2017matterport3d,
  title={{Matterport3D: Learning from RGB-D Data in Indoor Environments}},
  author={Angel Chang and Angela Dai and Thomas  Funkhouser and Maciej Halber and Matthias Nie{\ss}ner and Manolis Savva and Shuran Song and Andy Zeng and Yinda Zhang},
  booktitle={3DV},
  year={2017},
  pages={667--676},
}

@inproceedings{Sigurdsson2023RRExBoTRR,
  title={{RREx-BoT: Remote Referring Expressions with a Bag of Tricks}},
  author={Gunnar Sigurdsson and Jesse Thomason and Gaurav Sukhatme and Robinson Piramuthu},
  booktitle={IROS},
  year={2023},
  pages={5203-5210}
}

@inproceedings{Yenamandra2023HomeRobotOM,
  title={{HomeRobot: Open-Vocabulary Mobile Manipulation}},
  author={Sriram Yenamandra and A. Ramachandran and Karmesh Yadav and Austin Wang and Mukul Khanna and Th{\'e}ophile Gervet and Tsung Yang and Vidhi Jain and Alexander Clegg and others},
  booktitle={CoRL},
  year={2023},
  pages={1975--2011},
}

@inproceedings{clip,
  title={{Learning Transferable Visual Models From Natural Language Supervision}},
  author={Alec Radford and Jong Kim and Chris Hallacy and Aditya Ramesh and Gabriel Goh and Sandhini Agarwal and Amanda Askell and Pamela Mishkin and others},
  booktitle={ICML},
  pages={8748--8763},
  year={2021}
}

@inproceedings{blip2,
  title={{BLIP-2: Bootstrapping Language-Image Pre-training with Frozen Image Encoders and Large Language Models}},
  author={Li, Junnan and Li, Dongxu and Savarese, Silvio and Hoi, Steven},
  booktitle={ICML},
pages={19730--19742},
  year={2023}
}

@inproceedings{beit3,
  title={{Image as a Foreign Language: BEIT Pretraining for Vision and Vision-Language Tasks}},
  author={Wang, Wenhui and Bao, Hangbo and Dong, Li and Bjorck, Johan and Peng, Zhiliang and Liu, Qiang and Aggarwal, Kriti and others},
  booktitle={CVPR},
  year={2023},
  pages={19175-19186},
}

@inproceedings{longclip,
  title={{Long-CLIP: Unlocking the Long-Text Capability of CLIP}},
  author={Beichen Zhang and Pan Zhang and Xiaoyi Dong and Yuhang Zang and Jiaqi Wang},
  booktitle={ECCV},
  year={2024},
  pages = "310--325",
  numpages = {16},
}

@INPROCEEDINGS{nlmap,
  author={Chen, Boyuan and Xia, Fei and Ichter, Brian and Rao, Kanishka and Gopalakrishnan, Keerthana and Ryoo, Michael and others},
  booktitle={ICRA}, 
  title={{Open-vocabulary Queryable Scene Representations for Real World Planning}}, 
  year={2023},
  pages={11509-11522},
}

@ARTICLE{relaxformer,
  author={Yashima, Daichi and Korekata, Ryosuke and Sugiura, Komei},
  journal={IEEE RA-L}, 
  title={{Open-Vocabulary Mobile Manipulation Based on Double Relaxed Contrastive Learning With Dense Labeling}}, 
  year={2025},
  volume={10},
  number={2},
  pages={1728-1735},
}

@article{Xie2024EmbodiedRAGGN,
  title={{Embodied-RAG: General Non-parametric Embodied Memory for Retrieval and Generation}},
  author={Xie, Quanting and Min, So and Ji, Pengliang and Yang, Yue and Zhang, Tianyi and Xu, Kedi and Bajaj, Aarav and others},
  journal={arXiv preprint arXiv:2409.18313},
  year={2024}
}

@article{wang2025navraggeneratinguserdemand,
      title={{NavRAG: Generating User Demand Instructions for Embodied Navigation through Retrieval-Augmented LLM}}, 
      author={Zihan Wang and Yaohui Zhu and Gim Lee and Yachun Fan},
      journal={arXiv preprint arXiv:2502.11142},
      year={2025},
}

\end{document}